**Title**:
**Predicting drug properties with parameter-free machine learning: Pareto-Optimal Embedded Modeling (POEM)**

**Authors:**
Andrew E. Brereton[1], Stephen MacKinnon[1], Zhaleh Safikhani[1,2], Shawn Reeves[1], Sana Alwash[1], Vijay Shahani[1], Andreas Windemuth[1*]

**Affiliations:**
[1] Cyclica Inc. 207 Queens Quay W Suite 420, Toronto, ON M5J 1A7, Canada
[2] Vector Institute for Artificial Intelligence, Toronto, Ontario, Canada
* Corresponding author email addresses: windemut@yahoo.com (A.W.)

**Abstract:**
The prediction of absorption, distribution, metabolism, excretion, and toxicity (ADMET) of small molecules from their molecular structure is a central problem in medicinal chemistry with great practical importance in drug discovery. Creating predictive models conventionally requires substantial trial-and-error for the selection of molecular representations, machine learning (ML) algorithms, and hyperparameter tuning. A generally applicable method that performs well on all datasets without tuning would be of great value but is currently lacking. Here, we describe Pareto-Optimal Embedded Modeling (POEM), a similarity-based method for predicting molecular properties. POEM is a non-parametric, supervised ML algorithm developed to generate reliable predictive models without need for optimization. POEM's predictive strength is obtained by combining multiple different representations of molecular structures in a context-specific manner, while maintaining low dimensionality. We benchmark POEM relative to industry-standard ML algorithms and published results across 17 classifications tasks. POEM performs well in all cases and reduces the risk of overfitting.

**Introduction:**

  Chemical activity predictions continue to present a longstanding challenge with practical importance during pharmaceutical research and development. In particular, predictive tasks that associate chemical structures to their activity are known as Quantitative Structure Activity Relationships (QSARs) (*1*). In modern drug design programs, QSARs are used for modeling specific target biological activities or broader pharmacokinetic behaviours, including the absorption, distribution, metabolism, excretion, and toxicity of drug candidate molecules, collectively referred to as *ADMET* (*2–4*). All machine learning (ML) approaches to predict chemical activity have three fundamental requirements: [1] a *library* of diverse reference molecules with a known property to predict (labeled examples), [2] a form of molecular representation and [3] a discriminative supervised learning algorithm. In practical drug discovery applications, multiple algorithms and molecular representations are explored and optimized on a trial-and-error basis (*5*), since their performance varies considerably from predictive task to predictive task. This process generates final models which can vary considerably, even between extremely similar predictive tasks, in a manner that is seemingly arbitrary and often difficult to interpret.

  Molecular representations are a key component for predictive modeling of chemical activity. Different representations can be more or less relevant to different predictive tasks, not unlike the context-dependent use of different molecular representations in a laboratory. For example, acetylsalicylic acid can be described by several different common names: IUPAC name, chemical formula ($C_9H_8O_4$), a simplified molecular-input line-entry system (SMILES) string (eg. "O=C(C)Oc1ccccc1C(=O)O") (*6*, *7*), or a drawing of a 2-D structure (eg. as in Figure 1). Each of these representations is valid, but describe acetylsalicylic acid with differing levels of information content. For example, Acetozone has the same chemical formula as acetylsalicylic acid, but not the same 2-D structure; this is consistent with the observation that the chemical formula is a higher entropy description of a molecule than a drawing of its 2-D structure. For most ML applications, the molecular representation is defined in a way that is much simpler for a computer to extract useful information from, such as a binary or numeric vector (eg. Figure 1E).

  Physicochemical molecular descriptors are one conceptually simple way to create such a vector: by measuring a series of known properties (eg. mass, number of heteroatoms, charge, etc.) (*8*). While physicochemical descriptors have favorable interpretability, these overly simplistic representations of a molecule can lead to poor predictive power or a host of other problems associated with incorrect descriptor selections (*9*). Thus, an approach relying on physicochemical molecular descriptors may be robust for a single predictive task, but often will not be generalizable.

  Alternatively, molecules may be converted into a vector format in a process called molecular fingerprinting (*10*). Unlike physicochemical molecule descriptors, fingerprints do not need to have any human-obvious relationship to the properties of the molecule; they are generated by following strict algorithmic rules that associate individual positions in the vector to the presence/absence of specific substructures and substructure relationships. A primary advantage gained by using fingerprints, as opposed to purely physicochemical molecular descriptors, is that they are usually much more generalizable; a fingerprint of a molecule will often be useful to some degree across many different problems. Unfortunately, this generality comes with a lack of specificity, and risks training a model on syntactic features of the representation ("noise") rather than true details of the structure (*11*).

Most fingerprinting methods do lose information upon encoding, resulting in vectors that tentatively represent several different molecules (*12*, *13*). Typically, the loss of information is coupled to specific advantages such as improved speed, lower memory usage, algorithmic advantages, or increased density of useful information. For instance, Daylight fingerprints can exist as long, variable length fingerprints, or they can be 'folded' into a fixed-length representation, gaining speed and memory efficiency at the cost of reversibility (*10*). In addition to losing information, compressed (folded) fingerprints are prone to introducing algorithmic noise, leading to potential false equivalencies. While reversibility may be desirable in some applications, its is not always necessary.

The relationship between molecular features and chemical activity is established during the training stages of supervised learning algorithms. Chemical activities are predicted using industry-standard ML approaches for supervised learning, such support vector machines (SVMs), random forest modeling, ridge classification and neural networks. Typically, each algorithm has its own specific set of *hyperparameters* that influence the performance of the trained models. The problem of knowing which hyperparameters to use is itself an interesting and non-trivial topic of research (*14*). In practice, the optimal combination of hyperparameter values also varies across tasks and must be calibrated through a trial-and-error based optimization process. This high computational cost of retraining models can be a deterrent for incorporating new training data at a later stage, although this drawback is partially mitigated in some neural network based approaches through recent advances in transfer learning (*15*, *16*). When a supervised learning algorithm successfully establishes relationships between molecular features and chemical activities, the resulting model can be used to predict the chemical activity of new, previously unseen molecules.

Model interpretability is another important factor in the selection of supervised learning algorithms. Some algorithms, for example decision trees, provide human interpretable justifications for the associations between molecular features and chemical activities. For other algorithms, such as neural networks, the reasons for any given prediction can often be extremely hard to determine. In practice, the "black box" nature of these models can inhibit trust and possibly reduce the real-world utility of the model. Understanding this rationale can lead to new strategies for reducing bias, overfitting, and possibly even improve theory (*17*). Ideally, each prediction generated by the model has a clear rationale behind it, which can be understood by the researcher using the model.

Unfortunately, the combination of variable context-dependent molecular representations, choice of supervised learning algorithms and hyperparameterization is typically inconsistent between predictive tasks. Any specific combination of these three factors optimized for a specific chemical activity model will not be applicable to predicting other activities. In practice, prediction of seemingly related chemical properties may end up requiring models with different molecular features, algorithms, and/or parameters following substantial development effort by expert chemoinformaticians.

One approach to addressing inconsistency in the choice of molecular fingerprints and tentatively boosting performance is to use multiple fingerprints simultaneously as features. It is clear that distinct molecular representations possess varying amounts of useful information relating to specific problems (*i.e.* each molecular representation discards information, but not every method discards *the same* information). So, two fingerprints of the same molecule produced by two different low-information fingerprint methods will share some redundant information about the structure, and contain some unique information. By combining molecular representations, a greater amount of "true" information can be captured from the complete structure of the molecule. A naive way to combine molecular representations is to append the vectors generated by each fingerprint directly onto each other, to create a new, longer fingerprint, of multiple types. This approach has at least two major disadvantages: [1] for some supervised learning algorithms, the computational costs may become prohibitive when concatenating multiple fingerprints, due to increased length of the vector; and [2], the concatenation approach is prone to the creatively named "curse of dimensionality" (*18–20*). This phenomenon occurs when the ratio of training data to features is low, and overfitting becomes not just possible, but likely. Other methods approach this issue by using voting schemes to try to weigh consensus between models built using different fingerprints (*21*), or by nesting the selection of a fingerprint with hyperparameter selection to empirically determine the most effective fingerprint (*22*).

In this report, we describe Pareto-Optimal Embedded Modelling (POEM), a novel supervised learning method that reliably creates accurate models for predicting drug activity based on multiple representations of molecular structure. POEM evades the *curse of dimensionality* by using pareto multi-front optimization (*23*, *24*) to massively shrink the number of dimensions that define molecular similarity, in a context-dependent manner. A Pareto optimization algorithm is a powerful general approach for identifying optimal solutions in cases where there are more than one metric to optimize, that may not always be in agreement with each other. It has found broad applications including protein structure minimization (*25*, *26*) and lead optimization in drug design (*27*). Effectively, the context-dependent dimensionality reduction introduced by POEM enables the use of multiple fingerprints to describe every molecule in the reference molecule library, without introducing a risk of overfitting. The specific use of a pareto-based approach to defining similarity ensures that all comparisons remain 'like-to-like' and avoids the need for heuristic transformations and weighting schemas. POEM also has a number of additional functional advantages: the rationale for predictions are each interpretable, the algorithm has no hyperparameters, and models can be easily updated with new labeled reference molecules. This approach was designed for the rapid generation of multiple predictive models, without the need for expert intervention. We demonstrate the generalizability and consistency of POEM across a broad range of predictive tasks by modeling 17 ADMET properties of interest to pharmaceutical drug development.

**Materials and Methods:**

*Strategy*

POEM uses a Pareto dominance definition to combine multiple definitions of molecular similarity, into a robust metric suitable for supervised learning tasks. The POEM method has four major steps: [1] fingerprinting, [2] embedding the known molecules into a limited context based on similarity, [3] calculating Pareto dominance relationships for the known molecules, and [4] converting these dominance relationships into similarity scores and predictions. A high-level flowchart describing this process and the outputs at each step is provided in Figure S1.

*Step 1: Fingerprinting*

**N** molecular fingerprints techniques are applied to the target molecule and the reference library of **M** labeled compounds. Step 1 results in a matrix of **M**x**N** fingerprints for reference library and a vector of **N** fingerprints for the reference molecule. For this study, ten diverse and widely-used fingerprints were chosen. Detailed parameters and references for these ten fingerprints are provided in TABLE S1.

*Step 2: Embedding Known Molecules*

The fingerprint representations of the target molecule are embedded onto the chemical landscape of the reference library via a non-reversible transformation. For each fingerprint, Tanimoto distances (28) (Figure 1E) are calculated between the target molecule and all reference molecules in the library. Step 2 results in a matrix of **M**x**N** distance values, centered on the target molecule.

*Step 3: Calculating Dominance Relationships*

Although each molecule in the reference library is represented by a set of **N** distances, the specific values of the distances are not directly comparable. A distance of 0.4 may represent a significant match for one fingerprint, but random noise for another. To resolve this issue, Pareto dominance relationships are used to establish an overall distance. Here, the target molecule is selected as the *ideal objective* for multi-front optimization and the set of fingerprints represents **N**-dimensional space (27, 29). Dominance relationships between molecules from the reference library defined on the basis of which molecule is *closer* to the target. One reference library molecule may be closer than another to the target for all **N** distances, or a subset of the **N** dimensions. When evaluating the dominance of one reference library molecule (**A**) to another labeled molecule (**B**) across all 10 distances, closer distances are assigned a value of 1, ties are assigned a value of 0.5, and further distances are assigned a value of 0. A comparison vector **AB** = [1, 0, 1, 0.5, 0, 0.5, 0.5, 1, 1, 1] would indicate that molecule A is more similar to the target molecule in five fingerprint representations, tied in three representations, and more dissimilar using two fingerprint representations. Step 3 results in an **MxM** symmetric matrix of dominance relationships.

In a naive Pareto scheme, a molecule would *dominate* another molecule all its distances were as close or closer to the target molecule, and at least one distance was closer (*26*). POEM relaxes the naive definition of dominance, allowing a molecule to claim dominance over another, even if <=10% of its distance comparisons remain further from the target. In practice, this relaxation yields more dominance relationships overall when a larger number of fingerprints is used. The added dominance relationships reduce the likelihood of ties, which helps establish a complete ranking of all molecules in the labeled library relative to the target. Sample code for evaluating these relationships is provided in Supplemental Pseudocode 1.

*Step 4: Calculating Fitness Scores and Final Prediction*

Labeled reference molecules are ranked by their similarity to the target molecule by converting dominance relationships to a single-value *fitness score*. For a given molecule, its fitness is defined as:

$$Fitness_{i} = MeanDominance_{i} \cdot \frac{(NumDominating_{i} + 0.05)}{(NumSubmitting_{i} + 0.05)}$$

This schema favors labeled reference molecules which compare favorably "on average" for all fingerprints (*MeanDominance*), which dominate many other molecules (*NumDominating*), and which are not being dominated by others (*NumSubmitting*). This approach favors labeled molecules that are "best-of-class" across all metrics of similarity to the target molecule. Sample code for ranking molecules is provided in Supplemental Pseudocode 2.

Labeled reference molecules are ranked according to their fitness scores and summed to provide a "total fitness" value. In practice, fitness values vary by orders of magnitude between the most similar and dissimilar molecules. In some cases, the top few molecules could contribute the vast majority of the weight towards the summed fitness value. Alternatively, contribution towards the total fitness value may be more broadly distributed across the reference molecule library. The relative contribution of each labeled reference molecule to the total fitness score is then used as a weight towards each observed class label. Weighted averages are treated as probabilities of the target having any given label. Sample code evaluating probabilities is provided in Supplemental Pseudocode 3. Step 4 results in a Length **M** fitness vector of similar molecules, which is used to assign probabilities to each label class.

*Benchmarking POEM to Standard Approaches with 17 ADMET property predictions*

POEM was benchmarked relative to five standard supervised learning algorithms, across 17 predictive tasks related to drug ADMET properties. All 17 ADMET datasets were taken from public sources and range between 522 and 6505 labeled reference molecules. Each molecule was represented by a SMILES string. RDKit release 2018.09.1 was used to parse, canonicalize, and featurize small molecules. Molecules that could not be automatically processed by RDKit and molecules that have both positive and negative data labels were excluded from each dataset. Redundant data points representing experimental replicates were also removed. The *Supplementary Text* contains references and descriptions for each dataset used in this study, including the total number of training examples for each label after dataset cleaning. Python's scikit-learn package v0.19.1 (*30*) was used to build models for each of the five standard supervised learning algorithms: Gradient Boosting Classifier, Random Forest, Ridge Classifier, Stochastic Gradient Descent Classifier, and Support Vector Machine, representing a range of industry-standard supervised classifier types, which are suitable for the dataset sizes in this benchmark study. Each of the five standard supervised learning algorithms was trained using a grid search strategy for hyperparameter optimization, with an added nested layer evaluating performance separately for each of the fingerprints listed in Table S1. In contrast, POEM has no hyperparameters and considers all fingerprints simultaneously. In addition, a Molecule Graph Convolution (31, 32) model was trained using the GraphConv model in the DeepChem python package (33), on all 17 datasets, to provide a comparison to state-of-the-art deep neural network methods. Results are reported both from a naive model with default parameters (2 convolution layers of size 64, one dense layer of 128, 75 features per atom, 10 training epochs), and a hyperparameter optimized model (parameters selected after 20 rounds of 5-fold cross-validated Bayesian optimization, for reference see Table S4). Finally, we also evaluated POEM models that were limited to using only individual fingerprints, to provide an added comparison to a non-consensus, similarity-based classifier approach.

*Cross validation of predictive models*

For each of the five standard supervised learning approaches, models were trained and evaluated with a five-fold cross-validation strategy, using 80% of the dataset for training and hyperparameter optimization and withholding 20% for blind performance evaluation. The same 80% / 20% split was used to create a *POEM test set*, which provides a direct comparison to the five standard algorithms. POEM is also amenable to full 'leave-one-out' cross validation due to the lack of a computationally-expensive training process. This corresponding *POEM Full Set* evaluation provides an indication of predictive robustness with respect to dataset size. Additionally, nested cluster validation was performed as in Mayr *et al.* (34), with the added restriction that the test set must contain at least one example of each class.

**Results:**

*POEM outperforms five standard supervised learning methods for predicting 17 ADMET properties*

Table 1 provides a comparison between POEM performance relative to the top performing fingerprint/algorithm combinations generated from standard supervised learning approaches, across 17 different ADMET tasks. Across all tasks, the *POEM Test Set* outperforms the standard supervised learning algorithms, as determined by ROC area under the curve (AUC) score, in some cases by more than 10% (Figure 2). Additionally POEM performs approximately as well, or better, than the GraphConv neural network (Figure 2). Validation scores associated with standard classifiers are consistently better than test scores (Figure 3), indicating some degree of overfitting. In contrast, the *POEM Full Set* scores are sometimes higher and sometimes lower than the *POEM Test Set* scores. Generally, similarity between these scores is an indication of predictive robustness and resistance to overfitting. A notable exception in both cases is the AR dataset, which shows the largest score disparity between the *POEM Full Set*, *POEM Test Set.* This disparity and relatively low predictive performance may indicate an insufficient representation of chemical space in the underlying dataset.

To assess predictive robustness with regards to the random test set selection, 100 different 80% / 20% testing splits were performed on the Blood Brain Barrier (BBB) and the Caco-2 Permeability (Caco2) datasets (Figure S2). ROC AUC was computed for all 100 using both POEM and the previously determined best model and hyperparameters for each approach (as reported in Table S2). Compared to the standard classifiers, the POEM performance is higher for each test set, and overall shows less variability.

To assess POEM's ability to generalize, a nested cluster validation approach was used. This approach provides insight into how well POEM can make predictions for molecules highly dissimilar to any known reference molecules. We see that POEM does generalize well, especially for certain datasets (eg. BBB in Figure S3), though in some other cases performance is poor on dissimilar molecules (eg. AR in Figure S3). This approach was also applied to the GraphConv neural network, with comparably good results (Figure S4).

We also compared POEM performance with literature-reported results for models trained on the same datasets. Kansen et al. report expert-optimized models for the Ames Mutagenicity dataset built using molecular descriptors and seven binary classification tools (*35*). The top performing model in that study was a support vector machine (SVM) with ROC AUC of 0.86, whereas the non-parametric POEM automatically generated comparable models with 0.87 AUC. This study shares three common datasets with AdmetSAR, a predictive tool that uses substructure-based descriptors and support vector machines (*36*). AdmetSAR reports five-fold cross-validation ROC AUC values for: Blood Brain Barrier (0.9517), Human Intestinal Absorption (0.9458), Caco-2 Permeability (0.8216). These models were generated using different combinations of three fingerprints and three binary classification algorithms (*36*). The Androgen Receptor (AR) activity dataset was taken from the Tox21 Challenge, a federal collaboration involving NIH, EPA and the FDA aimed to develop better toxicity assessment methods. For this sub-challenge, 31 teams contributed different predictive models, with the leading ROC AUC at 0.828 (*37*). Consistently, POEM matched or outperformed expert-developed models reported in the literature.

We also compared the standard POEM approach to a modified variation that uses individual fingerprints rather than a consensus of ten, reporting all *leave-one-out* cross validation results for each dataset in Table S3. As expected, the standard consensus approach performs reliably well, appearing among the top models for each task and clearly outperforming any individual single-fingerprint across all tasks. The use of multiple fingerprints consensus also helps establish meaningful confidence scores associated with the predictive tasks. Figure 4 presents the distribution of POEM-predicted probability for the 'correct' label for the Caco-2 permeability dataset, across the consensus 10FP POEM, and each of the ten single-fingerprint models. In this figure, data points that lie below 0.5 represent an incorrect prediction and points closer to 100% and 0% represent higher confidence predictions. The Caco-2 dataset was chosen as an example dataset to demonstrate this principle, since the consensus approach ranked lower than three other single-fingerprint models. Performance varied across all fingerprinting methods, and was of approximately average predictive power overall. This figure demonstrates that the consensus model is better able to capture the confidence of a given prediction, as most of the observed incorrect predictions were made with lower confidence than for the single-fingerprint models. The single-fingerprint models almost exclusively produce highly confident correct predictions, or highly confident incorrect predictions.

**Discussion:**

At its core, POEM is a method based on measuring similarity, conceptually similar to a K-nearest neighbor approach (*20*). The predictive power of the method is based on the assumption that molecules with similar structures have similar properties. This approach has an established tradition (*38*), and is not in itself novel. POEM differs from these other approaches by intentionally restricting comparisons to relative similarity (through the embedding stage). Information is lost during the transformation of quantitative distances to relative similarity, such that similarity relationships evaluated are only meaningful in the context of the specific target molecule and predictive task. Specifically, the magnitude of any given distance has variable significance across fingerprints and predictive tasks. By ignoring the quantitative distribution of Tanimoto distances and operating only on less/greater comparisons between like fingerprints, POEM's treatment of reference molecule data is statistically *non-parametric*. The transformation to a 'distribution-free' representation of distances sidesteps the need for fingerprint-related transformation functions, weights or voting schemas. In turn, the entire landscape of labeled reference molecules can be used in making each prediction in a consistent and systematic manner, without the need to introduce fingerprint-related hyper-parameters to optimize from task-to-task. This approach leads to fast and objective model building, two major functional advantages of the algorithm.

While POEM's treatment of input data is statistically non-parametric, modification to the algorithm itself may impact performance. For example, adding new fingerprints may further improve performance. Future studies on larger datasets may identify new such modifications that produce significantly different optimal configurations from task-to-task, leading to optional hyper-parameters and a potential for optimization. Nonetheless, the results in this study demonstrate consistent performance using a static algorithm configuration and fingerprint selection. In this context, POEM is a powerful general purpose supervised machine learning approach that does not require hyperparameter optimization. POEM was designed to reduce the need for highly-tuned models, crafted by ML experts. Conceptually, the POEM algorithm is also applicable to problems outside of chemistry, as long as the object of the prediction has multiple representations which have metrics to define similarity.

We have shown that the increased performance and consistency of POEM is attributed to the use of multiple fingerprints simultaneously, as previously observed in other similarity approaches such as the *Similarity Ensemble Approach* (*38*). POEM models created using 10 fingerprints outperform those created by individual fingerprints with few exceptions, which may be attributed to random variation. This is further supported by the observation that the optimal fingerprint differs on a task-by-task basis when evaluating the standard supervised learning methods (Table 1) or single fingerprint POEM models (Table S3). Even seemingly related tasks, such as Cytochrome P450 activity predictions, are best addressed with different fingerprints for different isoforms. If algorithmic noise is responsible for variation between models generated using different fingerprints or hyperparameters, then model performance may be compromised in subsequent real world applications. The use of multiple fingerprints simultaneously in POEM side-steps this issue, providing reliable performance across a range of predictive tasks. Efforts were initially made to also build predictive models for benchmarking using concatenated fingerprints, but computational runtime was deemed cost-prohibitive, demonstrating instead a distinct speed advantage to the POEM strategy for combining representations.

POEM is also seen to generalize well, in terms of the ability to make predictions for molecules unlike any known reference molecules (Figure S3). More specifically, we observe that it performs as well or better than the GraphConv neural network (Figure S4). This network in particular was chosen because it is known to be highly performative (31), and was also not too cost-prohibitive (though we were forced to limit the amount of nested cluster validation to two properties due to compute cost concerns).

While this study limits the scope of POEM to classification problems, the fundamental relationship between molecular similarity and activity established by POEM may provide a suitable framework for developing regression models. Preliminary findings applying POEM to four standard benchmark datasets presented in Supplemental Figure 5 demonstrate favorable performance over leading deep learning frameworks to develop regression models for chemical activity. Future studies using a broader range of datasets would provide a broader understanding on POEM's utility towards regression problems.

We have identified a number of functional advantages to POEM, including model building speed, reliability, predictive power, objectivity, ease of use, and model interpretability. POEM is particularly well suited to applications requiring automation due to its objective nature and reduced risk of overfitting. Highly automated applications may include: models built upon large-scale data mining expeditions, datasets with frequent updates, or model building by subject-area field experts without first-hand experience developing ML models. The similarity-based nature of this algorithm helps provide model interpretability, as each prediction is coupled with the list of reference molecules, their relative similarity, and their labels.

The above notwithstanding, there are important trade-offs associated with the POEM approach. Mainly, POEM has high algorithmic complexity associated with the generation of an **MxM** dominance matrix. Due to POEM's instance-based learning nature, predicting each unlabeled molecule scales proportionally with the square of the number of molecules in the reference molecule library. Algorithmically, this prediction stage is significantly slower when compared to other standard supervised learning methods. In practice however, POEM can handle datasets up to 100,000 training examples on modern personal computers (4 Core CPU, 16Gb RAM, Ubuntu 18.04). Applying POEM to ligand-based virtual screening using libraries with millions of molecules may also impose a technical challenge, requiring distributed computing solutions. Future heuristic approximations may improve POEM dataset scalability, including batching the dominance calculations (which would move POEM to O(nlogn) time rather than O(n^2), at the cost of completely sorting all reference molecules). As it stands, we have not made these changes at present, as datasets of fewer than 10,000 reference molecules are still capable of performing predictions in the range of milliseconds to seconds. In our experience, these timescales are suitable for most practical applications in drug discovery. Nonetheless, even with regard to larger datasets, the lack of dedicated 'training' and 'optimization' stages make up for limitations in speed in applications where fewer predictions need to be made. For instance, POEM models can be easily improved over time, without added computational cost, simply by adding new data into the set of labeled reference molecules. Additionally, we have observed that highly unbalanced datasets can behave poorly when using POEM to make predictions, and additional dataset balancing might be desirable for producing highly performant models.

As is always the case with machine learning approaches, the main determinant for predictive performance is the nature and quality of the data used for training. This is observed here in the contrast between good models (BBB, ERa and HIA) and bad models (Carcin, CYP), especially when looking at generalizability (Figure S2). Unfortunately, in the world of drug-property prediction, many of the best datasets are privately held, despite exciting but limited recent efforts to make some data available to researchers and the public (*39*).

In spite of the above limitations, we consider POEM to be a valuable addition to the roster of methods for supervised learning available today, especially given its lack of hyperparameters, high generalizability, and low cost.

**References:**

1. P. N. Craig, QSAR—Origins and Present Status: A Historical Perspective. *Drug Inf. J.* **18**, 123–130 (1984).
2. J. M. Sanders, D. C. Beshore, J. C. Culberson, J. I. Fells, J. E. Imbriglio, H. Gunaydin, A. M. Haidle, M. Labroli, B. E. Mattioni, N. Sciammetta, W. D. Shipe, R. P. Sheridan, L. M. Suen,

A. Verras, A. Walji, E. M. Joshi, T. Bueters, Informing the Selection of Screening Hit Series with in Silico Absorption, Distribution, Metabolism, Excretion, and Toxicity Profiles. *J. Med. Chem.* **60**, 6771–6780 (2017).
3. Clark, Pickett, Computational methods for the prediction of "drug-likeness." *Drug Discov. Today*. **5**, 49–58 (2000).
4. M. J. Waring, J. Arrowsmith, A. R. Leach, P. D. Leeson, S. Mandrell, R. M. Owen, G. Pairaudeau, W. D. Pennie, S. D. Pickett, J. Wang, O. Wallace, A. Weir, An analysis of the attrition of drug candidates from four major pharmaceutical companies. *Nat. Rev. Drug Discov.* **14**, 475–486 (2015).
5. Z. Wu, B. Ramsundar, E. N. Feinberg, J. Gomes, C. Geniesse, A. S. Pappu, K. Leswing, V. Pande, MoleculeNet: a benchmark for molecular machine learning. *Chem. Sci.* **9**, 513–530 (2018).
6. OpenSMILES specification, (available at http://opensmiles.org/opensmiles.html).
7. N. M. O'Boyle, Towards a Universal SMILES representation - A standard method to generate canonical SMILES based on the InChI. *J. Cheminformatics*. **4**, 22 (2012).
8. C. A. Lipinski, F. Lombardo, B. W. Dominy, P. J. Feeney, Experimental and computational approaches to estimate solubility and permeability in drug discovery and development settings. *Adv. Drug Deliv. Rev.* **46**, 3–26 (2001).
9. R. Gómez-Bombarelli, J. N. Wei, D. Duvenaud, J. M. Hernández-Lobato, B. Sánchez-Lengeling, D. Sheberla, J. Aguilera-Iparraguirre, T. D. Hirzel, R. P. Adams, A. Aspuru-Guzik, Automatic Chemical Design Using a Data-Driven Continuous Representation of Molecules. *ACS Cent. Sci.* **4**, 268–276 (2018).
10. Daylight Theory Manual, (available at http://www.daylight.com/dayhtml/doc/theory/).
11. R. Winter, F. Montanari, F. Noé, D.-A. Clevert, Learning continuous and data-driven molecular descriptors by translating equivalent chemical representations. *Chem. Sci.* **10**, 1692–1701 (2019).
12. N. M. O'Boyle, R. A. Sayle, Comparing structural fingerprints using a literature-based similarity benchmark. *J. Cheminformatics*. **8** (2016).
13. S. Riniker, G. A. Landrum, Open-source platform to benchmark fingerprints for ligand-based virtual screening. *J. Cheminformatics*. **5**, 26 (2013).
14. J. S. Bergstra, R. Bardenet, Y. Bengio, B. Kégl, in *Advances in Neural Information Processing Systems 24*, J. Shawe-Taylor, R. S. Zemel, P. L. Bartlett, F. Pereira, K. Q. Weinberger, Eds. (2011), pp. 2546–2554.
15. S. J. Pan, Q. Yang, A Survey on Transfer Learning. *IEEE Trans. Knowl. Data Eng.* **22**, 1345–1359 (2010).
16. R. Raina, A. Battle, H. Lee, B. Packer, A. Y. Ng, in *Proceedings of the 24th International Conference on Machine Learning* (2007), pp. 759–766.
17. A. Mordvintsev, C. Olah, M. Tyka, Deepdream-a code example for visualizing neural networks. *Google Res.* **2** (2015).
18. C. C. Aggarwal, in *Proceedings of the 31st International Conference on Very Large Data Bases* (2005), pp. 901–909.
19. J. H. Friedman, On Bias, Variance, 0/1—Loss, and the Curse-of-Dimensionality. *Data Min. Knowl. Discov.* **1**, 55–77 (1997).
20. P. Indyk, R. Motwani, in *Proceedings of the Thirtieth Annual ACM Symposium on Theory of Computing* (1998), pp. 604–613.
21. Z. Wang, L. Liang, Z. Yin, J. Lin, Improving chemical similarity ensemble approach in target prediction. *J. Cheminformatics*. **8**, 20 (2016).
22. S. L. Dixon, J. Duan, E. Smith, C. D. Von Bargen, W. Sherman, M. P. Repasky, AutoQSAR:

an automated machine learning tool for best-practice quantitative structure-activity relationship modeling. *Future Med. Chem.* **8**, 1825–1839 (2016).
23. M. Magill, M. Quinzii, *Theory of Incomplete Markets* (MIT Press, 2002).
24. P. Ngatchou, A. Zarei, A. El-Sharkawi, in *Proceedings of the 13th International Conference on, Intelligent Systems Application to Power Systems* (2005), pp. 84–91.
25. Y. Li, A. Yaseen, in *AAAI Workshop* (2013), pp. 32–37.
26. Y. Li, I. Rata, E. Jakobsson, Sampling Multiple Scoring Functions Can Improve Protein Loop Structure Prediction Accuracy. *J. Chem. Inf. Model.* **51**, 1656–1666 (2011).
27. J. Besnard, G. F. Ruda, V. Setola, K. Abecassis, R. M. Rodriguiz, X.-P. Huang, S. Norval, M. F. Sassano, A. I. Shin, L. A. Webster, F. R. C. Simeons, L. Stojanovski, A. Prat, N. G. Seidah, D. B. Constam, G. R. Bickerton, K. D. Read, W. C. Wetsel, I. H. Gilbert, B. L. Roth, A. L. Hopkins, Automated design of ligands to polypharmacological profiles. *Nature*. **492**, 215–220 (2012).
28. D. Bajusz, A. Rácz, K. Héberger, Why is Tanimoto index an appropriate choice for fingerprint-based similarity calculations? *J. Cheminformatics*. **7**, 20 (2015).
29. W. Zhang, J. Pei, L. Lai, Computational Multitarget Drug Design. *J. Chem. Inf. Model.* **57**, 403–412 (2017).
30. F. Pedregosa, G. Varoquaux, A. Gramfort, V. Michel, B. Thirion, O. Grisel, M. Blondel, P. Prettenhofer, R. Weiss, V. Dubourg, others, Scikit-learn: Machine learning in Python. *J. Mach. Learn. Res.* **12**, 2825–2830 (2011).
31. D. Duvenaud, D. Maclaurin, J. Aguilera-Iparraguirre, R.l Gómez-Bombarelli, T. Hirzel, A. Aspuru-Guzik, R. P. Adams, Convolutional Networks on Graphs for Learning Molecular Fingerprints. ArXiv:1509.09292, (2015).
32. S. Kearnes, K. McCloskey, M. Berndl, V. Pande, P. Riley, Molecular graph convolutions: moving beyond fingerprints. J Comput Aided Mol Des 30, 595–608. (2016)
33. B. Ramsundar, P. Eastman, P. Walters, V. Pande, K. Leswing, Z. Wu, Deep Learning for the Life Sciences, O'Reilly Media, (2019)
34. A. Mayr, G. Klambauer, T. Unterthiner, M. Steijaert, J. K. Wegner, H. Ceulemans, D.-A. Clevert, S. Hochreiter, Large-Scale Comparison of Machine Learning Methods for Drug Target Prediction on ChEMBL, *Chem. Sci.* **9**, 5441–5451 (2018)
35. K. Hansen, S. Mika, T. Schroeter, A. Sutter, A. Ter Laak, T. Steger-Hartmann, N. Heinrich, K.-R. Müller, Benchmark data set for in silico prediction of Ames mutagenicity. *J. Chem. Inf. Model.* **49**, 2077–2081 (2009).
36. F. Cheng, W. Li, Y. Zhou, J. Shen, Z. Wu, G. Liu, P. W. Lee, Y. Tang, admetSAR: a comprehensive source and free tool for assessment of chemical ADMET properties. *J. Chem. Inf. Model.* **52**, 3099–3105 (2012).
37. R. Huang, M. Xia, D.-T. Nguyen, T. Zhao, S. Sakamuru, J. Zhao, S. A. Shahane, A. Rossoshek, A. Simeonov, Tox21Challenge to build predictive models of nuclear receptor and stress response pathways as mediated by exposure to environmental chemicals and drugs. *Front. Environ. Sci.* **3**, 85 (2016).
38. M. J. Keiser, B. L. Roth, B. N. Armbruster, P. Ernsberger, J. J. Irwin, B. K. Shoichet, Relating protein pharmacology by ligand chemistry. *Nat. Biotechnol.* **25**, 197–206 (2007).
39. S. Ekins, A. J. Williams, When pharmaceutical companies publish large datasets: an abundance of riches or fool's gold? *Drug Discov. Today*. **15**, 812–815 (2010).

**Acknowledgments:**

We thank Lenny Morayniss and Joseph C. Somody for challenging discussions and feedback. We thank the RDKit open source project and scikit-learn for significant contributions to the scientific community and much useful code. We thank James Crompton and Marc Laforet for assisting with the benchmark standard classifiers and hyperparameter grid searches. We thank Naheed Kurji, CEO of Cyclica, for his unwavering support of this work without which it would not have been possible.

**Funding:** This work is partially supported through the Ontario Centre of Excellence TalentEdge Data Analytics Internship.
**Author contributions:** The method proposed in this paper was conceived by A.E.B, and developed by A.E.B. and A.W.; writing was done by A.E.B.; data collection and experiments were carried out by A.E.B., Z.S, S.A, S.R., and S.M.; and all authors were involved in planning, discussion, and preparation of the manuscript.
**Competing interests:** The algorithms publicly disclosed in this investigation were developed and commercialized by Cyclica Inc.
**Data and materials availability:** All data are publicly available, with information about data location reported in the supplementary information. POEM is available commercially through Cyclica's Ligand Express platform, and free of charge under the Cyclica Academic Partnership Program (CAPP), upon request.

**Figures:**

**Figure 1. Multiple representations of acetylsalicylic acid with varying levels of information. (A)** a 2D representation that explicitly shows all atoms and bonds within ASA, **(B)** a 2D line-diagram of ASA with details removed but isomorphic to the previous diagram, **(C)** a 3D depiction of ASA which highlights the geometry between each atom, and **(D)** a 3D space-filled depiction of ASA, which gives a sense of relative atom sizes. **(E)** Simplified depiction of a hypothetical molecular fingerprint showing two molecules converted into unique bits that can be directly compared (e.g. bits are used to calculate Tanimoto similarity(*28*) or distance (1 - Tanimoto similarity))

**Figure 2. Comparison of POEM performance versus industry-standard classifiers and literature reported values.** ROC area under the curve (AUC) scores for POEM and industry-standard ML methods tasked with predicting 17 ADMET properties are compared. Shown are the ROC AUC scores for full leave-one-out prediction with POEM (POEM full set, black circles), a 20% test set with POEM (POEM test set, black triangles), and the same 20% test set for best performing fingerprint and hyperparameters for each industry-standard ML method (colored straight lines). In addition, reported scores from AdmetSAR (*36*) (gray squares), an ADMET prediction tool, other Literature Values (*31*, *33*) (grey ovals), and a naive (gray diamonds) and hyperparameter optimized (gray rhomboids) GraphConv (*31*) neural network are also shown when applicable.

**Figure 3. Comparison of ROC AUC scores on test data versus leave-one-out cross validation scores (for POEM) or pre-validation scores (for industry-standard classifiers).** **(A)** ROC AUC scores are shown (y-axis) for each best performing classifier, hyperparameter, and fingerprint combination, for each of the 17 properties, for the same 20% test set as used in the POEM comparisons, and the mean of the 5-fold cross-validation score (Pre-Validation ROC AUC, x-axis), as assessed during model training. A dotted line is shown as a reference along the diagonal. **(B)** Shown are the POEM ROC AUC scores for each of the 17 properties being predicted, for a 20% test set (y-axis), and for leave-one-out cross validation on the remaining 80% set (the same set as in **A**) (x-axis). A dotted line is shown as a reference along the diagonal.

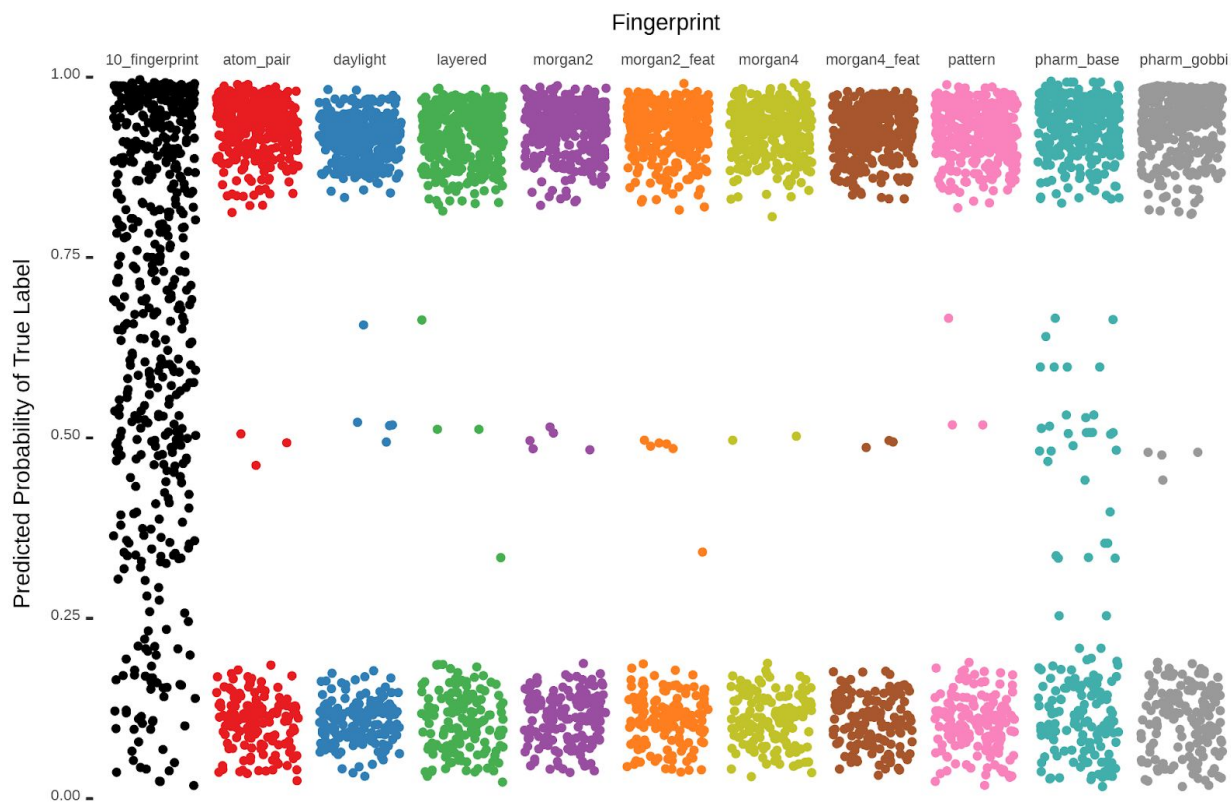

**Figure 4. Comparison of individual leave-one-out POEM predictions of Caco-2 permeability using 10-fingerprint consensus and 10 single fingerprints individually.** Each point represents the POEM prediction probability of the correct label for a reference labelled molecule withdrawn from the training data for a leave-one-out complete cross-validation. A probability of 1.0 represents 100% certainty of the correct prediction, and 0.0 represents 100% certainty of the incorrect prediction.

**Tables:**
**Table 1.** A summary of the performance (ROC AUC score) of 17 ADMET properties of POEM and best-performing, standard classifier-fingerprint combination. POEM Test: 80%/20% testing split for a direct comparison to traditional approaches, POEM Full: 'leave-one-out' full cross-validation.

| Property | POEM Full (ROC AUC) | POEM Test (ROC AUC) | Best Traditional (ROC AUC) | Best Classifier | Best Fingerprint |
|---|---|---|---|---|---|
| AMES Toxicity | 0.872 | 0.869 | 0.802 | Gradient Boosting | Pattern |
| Androgen Receptor | 0.864 | 0.744 | 0.687 | Gradient Boosting | Layered |
| Blood Brain Barrier | 0.979 | 0.981 | 0.916 | Stochastic Gradient Descent | Pattern |
| Caco-2 permeability | 0.828 | 0.822 | 0.726 | Gradient Boosting | Pharm Gobbi |
| Carcinogenic | 0.726 | 0.659 | 0.678 | Gradient Boosting | Layered |
| CYP450 2C9 Inhibitor | 0.801 | 0.840 | 0.594 | Stochastic Gradient Descent | Pharm Base |
| CYP450 2C9 Substrate | 0.690 | 0.612 | 0.507 | Stochastic Gradient Descent | Morgan Rad:4 |
| CYP450 2D6 Inhibitor | 0.750 | 0.718 | 0.566 | Stochastic Gradient Descent | Pharm Base |
| CYP450 2D6 Substrate | 0.778 | 0.688 | 0.594 | Ridge | Morgan Rad:2 |

| Property | | | | Model | Descriptor |
|---|---|---|---|---|---|
| CYP450 3A4 Inhibitor | 0.700 | 0.677 | 0.613 | Gradient Boosting | Pattern |
| CYP450 3A4 Substrate | 0.676 | 0.586 | 0.532 | Gradient Boosting | Morgan Rad:4 |
| Estrogen Receptor alpha | 0.963 | 0.961 | 0.855 | Gradient Boosting | Pharm Base |
| Human Intestinal Absorption | 0.949 | 0.954 | 0.883 | Stochastic Gradient Descent | Layered |
| Human Oral Bioavailability | 0.770 | 0.763 | 0.676 | Gradient Boosting | Pharm Base |
| Human Pregnane X Receptor | 0.895 | 0.876 | 0.646 | Ridge | Layered |
| P-glycoprotein Inhibitor | 0.945 | 0.954 | 0.885 | Gradient Boosting | Pattern |
| P-glycoprotein Recognition | 0.955 | 0.943 | 0.855 | Ridge | Morgan Rad:2 |

**Inventory of Supplementary Materials For:** *Predicting drug properties with parameter-free machine learning: Pareto-Optimal Embedded Modeling (POEM)*

Supplemental Methods:
- Supplemental Pseudocode 1. Evaluating dominance relationships
- Supplemental Pseudocode 2. Evaluating absolute distance rankings to target molecule
- Supplemental Pseudocode 3. Evaluating the probability of each label
- Figure S1. Flowchart Outlining the POEM Algorithm
- Table S1. List of 10 fingerprints used by POEM, using the RDKit implementation

Supplemental Results:
- Table S2. List of best-performing classifier, hyperparameter, and fingerprint
- Table S3: Non-consensus POEM evaluated on each fingerprint and task separately
- Table S4: ROC AUC scores and optimized GraphConv hyper-parameters
- Figure S2. Comparison of Blood Brain Barrier (BBB) and Caco-2 Permeability (Caco2) ROC AUC score distributions
- Figure S3. Nested cluster validation of POEM models for 17 ADMET properties
- Figure S4. Comparison of Blood Brain Barrier (BBB) and Caco-2 Permeability (Caco2) ROC AUC score distributions for POEM and GraphConv models using nested cluster validation
- Figure S5. Preliminary Observations of Regression Performance of POEM compared to GraphConv

Supplemental References For Datasets:
- BBB - Blood Brain Barrier
- Caco2 - Model for Intestinal Absorption
- HIA - Human Intestinal Absorption Human
- HOB - Human Oral Bioavailability
- PgpI - Inhibitor of p-glycoprotein 1
- PgpR - Substrate of p-glycoprotein 1
- CYP2C9I - Inhibitor of CYP2C9
- CYP2C9S - Substrate of CYP2C9
- CYP2D6I - Inhibitor of CYP2D6
- CYP2D6S - Substrate of CYP2D6
- CYP3A4I - Inhibitor of CYP3A4
- CYP3A4S - Substrate of CYP3A4
- AMES - A Mutagenicity Assay
- AR - Androgen Receptor Agonist
- Carcin - Carcinogenic potency
- ERa - Estrogen receptor 1 Agonist
- hPXR - Human pregnane X receptor agonist
- ESOL
- Freesolv
- Lipholicity
- LogS

# Supplemental Methods

## *Supplemental Pseudocode 1: Evaluating the dominance relationships*

```
"""
calculate the dominance relationships between all the children in the set
this refers to how much "closer" a child is to the template molecules, according to what numbers of methods
:param vecs: The distances of one molecule from all training cases
:param weights: array of weights for each fingerprint from 0.0-1.0 of how good the fp is
:return: dominance matrix, dominant count, subordinate count.
"""

M : Number of molecules in training set
N: Number of fingerprint types

dom_matrix[i,j] : an MxM matrix that will contain the dominance scores.
dom[i] : a vector of length M, listing the number of molecules dominated by each molecule i
sub[i] : a vector of length M, listing the number of molecules that dominate each molecule i
dist[k, i] : an MxN matrix of Tanimoto distances between the test molecule and molecule i, using fingerprint k

# precompute scores
score = [0.5 for k in 0 .. N]

for i in 0 .. M:
        s = [0, 0, ..., 0] x M
        num_better = [0, 0, ...., 0] x M
        num_worse  = [0, 0, ...., 0] x M
        for k in 0 .. N:
                v = dist[k,i]
                c = 0.5 + sign(dist[k] - v) * score[k]
                s = s + c
                num_better = num_better + (1 where c > 0.75, 0 otherwise)
                num_worse  = num_worse + (1 where c < 0.25, 0 otherwise)
        num_tied = N - (num_better + num_worse)
        dom_check = ((num_better + num_tied) / N) > 0.9
        sub_check = ((num_worse  + num_tied) / N) > 0.9

        dom_matrix[i] = s / N
        dom[i] += count(dom_check and not sub_check)
        sub[i] += count(sub_check and not dom_check)

return dom_matrix, dom, sub
```

## *Supplemental Pseudocode 2: Evaluating absolute distance rankings to target molecule*

```
"""
calculate the final rankings, based on the dominance matrix/information
"""
fitness[i] : M dimensional vector that will contain the fitness values

for i in 0 .. M:
        # the fitness is the sum of the dominance scores for each child
        s = sum(dom_matrix[i])
        # then, weight the score by the number of children dominating it
        # this favors non-dominated children
        fitness[i] = (float(s)) * (dom[i] + 0.05)) / (sub[i] + 0.05)
return fitness
```

## *Supplemental Pseudocode 3. Evaluating the probability of each class label*

```
"""
```

Compute the prediction probabilities
"""

Np : Number of property values (usually 2)

fitness[k,i] : fitness vector for each property value k, as above.

hit = [0, ..., 0] x Np  # compute all property values at once
for i in 0 .. N:
            j = poem.molProp[i] # Value of property for molecule i
          hit[j] += fitness[j][i]
probs = hit / sum(hit)  # Normalize probability

# Predicted value is that with maximum probability in 'probs'

**Figure S1. Flowchart Outlining the POEM Algorithm**

**Table S1. List of 10 fingerprints used by POEM, using the RDKit implementation**

| Fingerprint | Parameters |
|---|---|
| AtomPair | N/A |
| RDKit Default (Daylight) | N/A |
| Layered | N/A |
| Morgan (Radius=2) | Radius=2, useChirality=True, useFeatures=False |
| Morgan (Radius=2) with Features | Radius=2, useChirality=True, useFeatures=True |
| Morgan (Radius=4) | Radius=4, useChirality=True, useFeatures=False |
| Morgan (Radius=4) with Features | Radius=4, useChirality=True, useFeatures=True |
| Pattern | N/A |
| PharmBase | sigFactory:<br>feat_defs = os.path.join(RDConfig.RDDataDir, 'BaseFeatures.fdef')<br>feat_factory = ChemicalFeatures.BuildFeatureFactory(feat_defs)<br>sig_factory = SigFactory(feat_factory, minPointCount=2, maxPointCount=3, trianglePruneBins=False)<br>sig_factory.SetBins([(0, 2), (2, 5), (5, 8)])<br>sig_factory.Init() |
| PharmGobbi | sigFactory: rdkit.Chem.Pharm2D.Gobbi_Pharm2D.factory |

## Supplemental Results

**Table S2. List of best-performing classifier, hyperparameter, and fingerprint for each of 17 ADMET properties identified by grid search.**

| Property | Classifier | Hyperparameters | Fingerprint |
|---|---|---|---|
| AMES Toxicity | Gradient Boosting | {'learning_rate': 0.08, 'loss': 'exponential', 'max_depth': 7} | Pattern |
| Androgen Receptor | Gradient Boosting | {'learning_rate': 0.1, 'loss': 'deviance', 'max_depth': 5} | Layered |
| Blood Brain Barrier | Stochastic Gradient Descent | {'penalty': 'l1'} | Pattern |
| Caco-2 permeability | Gradient Boosting | {'learning_rate': 0.1, 'loss': 'exponential', 'max_depth': 5} | Pharm Gobbi |
| Carcinogenic | Gradient Boosting | {'learning_rate': 0.08, 'loss': 'exponential', 'max_depth': 7} | Layered |
| CYP450 2C9 Inhibitor | Stochastic Gradient Descent | {'penalty': 'elasticnet'} | Pharm Base |
| CYP450 2C9 Substrate | Stochastic Gradient Descent | {'penalty': 'l1'} | Morgan Rad:4 |
| CYP450 2D6 Inhibitor | Stochastic Gradient Descent | {'penalty': 'l1'} | Pharm Base |
| CYP450 2D6 Substrate | Ridge | {'alpha': 5} | Morgan Rad:2 |
| CYP450 3A4 Inhibitor | Gradient Boosting | {'learning_rate': 0.08, 'loss': 'deviance', 'max_depth': 7} | Pattern |
| CYP450 3A4 Substrate | Gradient Boosting | {'learning_rate': 0.1, 'loss': 'exponential', 'max_depth': 7} | Morgan Rad:4 |
| Estrogen Receptor alpha | Gradient Boosting | {'learning_rate': 0.1, 'loss': 'exponential', 'max_depth': 5} | Pharm Base |
| Human Intestinal Absorption | Stochastic Gradient Descent | {'penalty': 'elasticnet'} | Layered |

| Human Oral Bioavailability | Gradient Boosting | {'learning_rate': 0.08, 'loss': 'exponential', 'max_depth': 5} | Pharm Base |
| --- | --- | --- | --- |
| Human Pregnane X Receptor | Ridge | {'alpha': 1} | Layered |
| P-glycoprotein Inhibitor | Gradient Boosting | {'learning_rate': 0.1, 'loss': 'exponential', 'max_depth': 5} | Pattern |
| P-glycoprotein Recognition | Ridge | {'alpha': 5} | Morgan Rad:2 |

*Table S3: Non-consensus POEM evaluated on each fingerprint and task separately.*
Performance is provided as the full leave-one-out cross validation ROC AUC score in each case.

| Dataset | 10FP | Atom Pair | RDKit | Layered | Morgan R2 | Morgan R4 | Morgan R2F | Morgan R4F | Pattern | Pharm Base | Pharm Gobbi |
| --- | --- | --- | --- | --- | --- | --- | --- | --- | --- | --- | --- |
| AMES Toxicity | **0.8718** | 0.8593 | 0.8565 | 0.8575 | 0.8605 | 0.8561 | 0.8701 | 0.8651 | 0.8445 | 0.7953 | 0.8097 |
| Androgen Receptor | 0.8637 | 0.8652 | 0.8391 | 0.8519 | 0.8602 | 0.8605 | 0.8522 | 0.8567 | 0.8672 | 0.8382 | **0.8684** |
| Blood Brain Barrier | **0.9789** | 0.9658 | 0.9553 | 0.9658 | 0.9694 | 0.9691 | 0.9662 | 0.9662 | 0.9630 | 0.9679 | 0.9728 |
| Caco-2 permeability | 0.8280 | 0.8286 | 0.7591 | 0.7891 | 0.8157 | 0.8170 | 0.8266 | **0.8300** | 0.7820 | 0.8049 | 0.8282 |
| Carcinogenic | 0.7265 | 0.7049 | 0.7180 | **0.7322** | 0.7272 | 0.7149 | 0.7229 | 0.7141 | 0.7253 | 0.6395 | 0.7027 |
| CYP450 2C9 Inhibitor | **0.8015** | 0.7543 | 0.7410 | 0.7733 | 0.7613 | 0.7680 | 0.7842 | 0.7859 | 0.7826 | 0.7819 | 0.7602 |
| CYP450 2C9 Substrate | 0.6902 | 0.6843 | 0.6454 | 0.6787 | **0.6965** | 0.6935 | 0.6665 | 0.6696 | 0.6823 | 0.6776 | 0.6328 |
| CYP450 2D6 Inhibitor | **0.7504** | 0.7393 | 0.6825 | 0.6974 | 0.7433 | 0.7373 | 0.7153 | 0.7137 | 0.7237 | 0.7126 | 0.7283 |
| CYP450 2D6 Substrate | 0.7780 | 0.7858 | 0.7342 | 0.7665 | **0.7860** | 0.7844 | 0.7679 | 0.7675 | 0.7502 | 0.7804 | 0.7838 |
| CYP450 3A4 Inhibitor | 0.6997 | 0.6897 | 0.6720 | 0.6500 | 0.6936 | 0.6968 | **0.7001** | 0.7008 | 0.7075 | 0.6676 | 0.6623 |
| CYP450 3A4 Substrate | 0.6759 | **0.6873** | 0.6437 | 0.6391 | 0.6776 | 0.6694 | 0.6768 | 0.6716 | 0.6548 | 0.6360 | 0.6512 |
| Estrogen Receptor alpha | **0.9630** | 0.9532 | 0.9437 | 0.9429 | 0.9579 | 0.9578 | 0.9619 | 0.9627 | 0.9422 | 0.9501 | 0.9501 |

| Property | | | | | | | | | | |
|---|---|---|---|---|---|---|---|---|---|---|
| Human Intestinal Absorption | 0.9488 | 0.9557 | 0.8777 | 0.9011 | **0.9560** | 0.9493 | 0.9532 | 0.9503 | 0.8807 | 0.8961 | 0.8904 |
| Human Oral Bioavailability | **0.7704** | 0.7627 | 0.7293 | 0.7191 | 0.7533 | 0.7490 | 0.7606 | 0.7637 | 0.7088 | 0.7353 | 0.7498 |
| Human Pregnane X Receptor | 0.8950 | 0.8996 | 0.7644 | 0.8637 | 0.8743 | 0.8781 | **0.9038** | 0.9013 | 0.8787 | 0.8475 | 0.8298 |
| P-glycoprotein Inhibitor | **0.9447** | 0.9348 | 0.9111 | 0.9094 | 0.9380 | 0.9371 | 0.9411 | 0.9424 | 0.9320 | 0.9310 | 0.9300 |
| P-glycoprotein Recognition | **0.9553** | 0.9349 | 0.8219 | 0.8929 | 0.9410 | 0.9395 | 0.9374 | 0.9373 | 0.9052 | 0.8896 | 0.9247 |

Table S4: ROC AUC scores and optimized GraphConv hyper-parameters for each property prediction task.

| Property | Naive AUC | Optimized AUC | Learning rate | Batch size | Gc Layer 1 | Gc Layer 2 | Dense layer | Number Atom features |
|---|---|---|---|---|---|---|---|---|
| AMES | 0.875 | **0.876** | 0.000183 | 70 | 220 | 120 | 300 | 75 |
| AR | **0.767** | 0.736 | 0.000103 | 70 | 70 | | 310 | 70 |
| BBB | 0.979 | **0.981** | 0.000184 | 10 | 500 | 230 | 350 | 75 |
| Caco2 | 0.798 | **0.810** | 0.003091 | 30 | 110 | 160 | 70 | 70 |
| Carcin | **0.692** | 0.688 | 0.044583 | 90 | 70 | | 430 | 70 |
| CYP2C9I | **0.695** | 0.685 | 0.010737 | 80 | 260 | | 170 | 75 |
| CYP2C9S | 0.564 | **0.607** | 0.006779 | 80 | 240 | | 150 | 75 |
| CYP2D6I | **0.743** | 0.741 | 0.000134 | 30 | 110 | | 180 | 80 |
| CYP2D6S | **0.744** | 0.700 | 0.000337 | 80 | 80 | 90 | 120 | 75 |
| CYP3A4I | **0.643** | 0.628 | 0.016852 | 30 | 360 | 90 | 100 | 75 |
| CYP3A4S | 0.609 | **0.622** | 0.022635 | 80 | 460 | | 440 | 80 |
| ERa | 0.957 | **0.958** | 0.000146 | 50 | 250 | 140 | 300 | 80 |
| HIA | 0.899 | **0.968** | 0.001435 | 10 | 110 | | 150 | 85 |
| HOB | 0.718 | **0.720** | 0.000103 | 70 | 180 | 410 | 60 | 75 |
| hPXR | 0.868 | **0.872** | 0.003037 | 20 | 260 | | 110 | 75 |
| PgpI | **0.939** | 0.933 | 0.072239 | 40 | 100 | | 140 | 75 |
| PgpR | 0.937 | **0.951** | 0.000151 | 10 | 130 | | 100 | 75 |

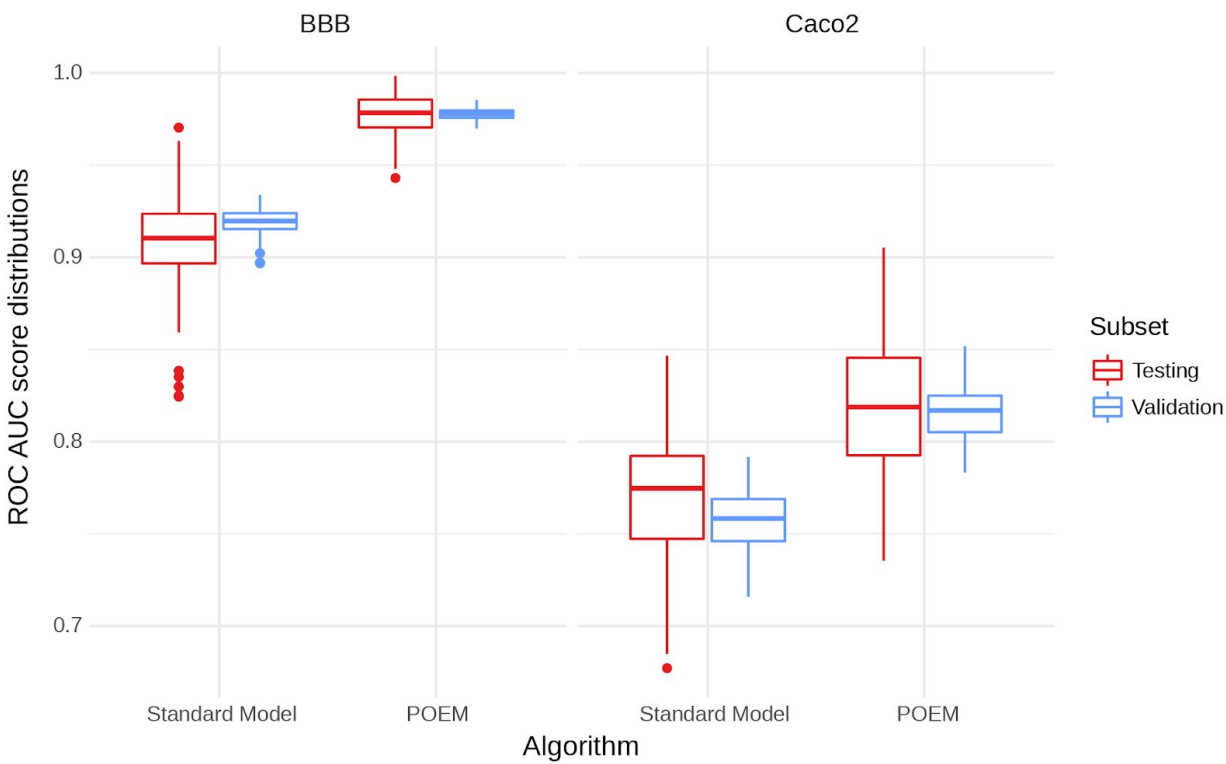

**Figure S2. Comparison of Blood Brain Barrier (BBB) and Caco-2 Permeability (Caco2) ROC AUC score distributions.** Shown are standard box plots of the distribution of ROC AUC scores from either a withheld 20% test-set (red) or an 80% 'training' set (blue) that represents the pre-validation scores for the industry-standard methods, or the leave-one-out cross-validation scores on the same reference molecule library when used by POEM. Model performance was calculated from 100 random splits, each split being used to generate both a POEM model, and an industry-standard model, for BBB and Caco2 (using their respective best method, fingerprint, and parameters, as listed in Table S2).

**Figure S3. Nested cluster validation of POEM models for 17 ADMET properties.** Shown are standard box plots of the distribution of ROC AUC scores from ten repeats of a nested cluster validation (34). Points are overlaid, and show the ROC AUC scores directly, colored to indicate the ratio of molecules in the test set relative to the 'training' set. The x-axis shows the minimum Tanimoto distance (calculated on Morgan R4) between any test set molecule and any training set molecule. As this distance increases, more molecules are removed from the training set, and predictions are being made on molecules that are increasingly unlike any reference molecules in the training set.

**Figure S4. Comparison of Blood Brain Barrier (BBB) and Caco-2 Permeability (Caco2) ROC AUC score distributions for POEM and GraphConv models using nested cluster validation.** Shown are standard box plots of the distribution of ROC AUC scores from ten repeats of a nested cluster validation (34). Points are overlaid, and show the ROC AUC scores directly. Scores are compared for Naive GraphConv (NGC) (red dots), hyperparameter optimized GraphConv (OGC) (blue dots), and POEM (green dots). The top panel label shows the minimum Tanimoto distance (calculated on Morgan R4) between any test set molecule and any training set molecule. As this distance increases, more molecules are removed from the training set, and predictions are being made on molecules that are increasingly unlike any reference molecules in the training set. The rightmost panel label indicates which property is being predicted, BBB, or Caco2.

**Figure S5. Preliminary Observations of Regression Performance of POEM compared to GraphConv.** Shown are standard box plots of the distribution of root mean squared error on the test set from ten repeats of a cluster validation (34). The errors are examined for 4 properties (each chart) and are compared for Naive GraphConv (NGC) (orange boxes), hyperparameter optimized GraphConv (OGC) (green boxes), and POEM (blue boxes). The x-axis shows the minimum permitted Tanimoto distance (calculated on Morgan R4) between any test set molecule and any training set molecule. As this distance increases, more molecules are removed from the training set, and predictions are being made on molecules that are increasingly unlike any reference molecules in the training set.

# Supplemental References for Datasets

## *Absorption and Distribution*
### BBB - Blood Brain Barrier
The blood brain barrier is an integral part of the central nervous system that protects the brain and spinal column for potentially hazardous material, including xenobiotic drugs. This model will predict whether or not a small molecule will likely traverse the complex filtering mechanism of the capillaries that carry blood to the brain and spinal cord tissue.
Dataset size: 1585 (1276 Passes BBB | 309 Does not pass BBB)
- Shen J., et al. Estimation of ADME Properties with Substructure Pattern Recognition. J. Chem. Inf. Model. 50, 1034-1041 (2010)

### Caco2 - Model for Intestinal Absorption
The Caco-2 cell monolayer is an established in vitro model that screens molecules for their intestinal absorption and efflux liabilities. This model predicts if query ligands will be permeable in a Caco-2 cell monolayer experiment. In this binary classifier, the intestinal permeability value of $8×10^{−6}$ cm/s is used to determine whether a query ligand is permeable
Dataset: 522 (231 Permeable | 291 Not permeable)
- The, H. P., et al. In Silico Prediction of Caco-2 Cell Permeability by a Classification QSAR Approach. Mol Inform. 30, 376-385 (2011)

### HIA - Human Intestinal Absorption Human
Intestinal Absorption is the process through which orally administered drugs are absorbed from the intestine into the bloodstream. This model predicts whether a query ligand will likely be absorbed in the gut. In this binary classifier, the fraction threshold of 30% is used to determine whether a query ligand is absorbed through the intestine.
Dataset: 578 (500 Absorbed | 78 Not absorbed)
- Shen J., et al. Estimation of ADME Properties with Substructure Pattern Recognition. J. Chem. Inf. Model. 50, 1034-1041 (2010)

### HOB - Human Oral Bioavailability
Human Oral Bioavailability is the fraction of an orally administered drug that is absorbed and reaches systemic circulation. In this binary classifier, the fraction threshold of 50% is used to determine whether a query ligand is orally bioavailable.
Dataset: 995 (509 orally bioavailable | 486 Not orally bioavailable)
- Kim M., et al. Critical Evaluation of Human Oral Bioavailability for Pharmaceutical Drugs by Using Various Cheminformatics Approaches. Pharm Res. 31, 1002-1014 (2014)

## *Excretion*
### PgpI - Inhibitor of p-glycoprotein 1

P-glycoprotein 1 is an ATP-dependent drug efflux pump for xenobiotic compounds with broad substrate specificity. It is responsible for decreased drug accumulation in multidrug-resistant cells and often mediates the development of resistance to anticancer drugs. This model will predict whether a query ligand will likely inhibit p-gp, resulting in an a predicted accumulation of xenobiotic compounds, a key consideration in drug safety and drug-drug interactions.
Dataset: 1216 (650 Inhibits Pgp | 566 Does not inhibit Pgp)
- Bocci G., et al. ADME-Space: a new tool for medicinal chemists to explore ADME properties. Scientific Reports. 7, 6359 (2017)

### PgpR - Substrate of p-glycoprotein 1
P-glycoprotein 1 is an ATP-dependent drug efflux pump for xenobiotic compounds with broad substrate specificity. It is responsible for decreased drug accumulation in multidrug-resistant cells and often mediates the development of resistance to anticancer drugs. This model will predict if a query ligand is a substrate of p-gp, causing the molecule to be rapidly cleared from the cell impacting the drug efficacy.
Dataset: 929 (445 Pgp substrate | 484 Not a Pgp substrate)
- Levatic J., et al. Accurate Models for P-gp Drug Recognition Induced from a Cancer Cell Line Cytotoxicity Screen. J. Med. Chem. 56, 5691-5708 (2013)

*Metabolism*
### CYP2C9I - Inhibitor of CYP2C9
Cytochrome P450 enzymes are essential for the metabolism of many medications. Although this class has more than 50 enzymes, six of them metabolize 90 percent of drugs, with the two most significant enzymes being CYP3A4 and CYP2D6. This model predicts whether the ligand will likely inhibit the activity of CYP2C9, an important consideration for drug-contraindications, efficacy, and overall drug safety.
Dataset: 691 (166 Yes | 525 No)
References:
Yap CW., et al. Prediction of cytochrome P450 3A4, 2D6, and 2C9 inhibitors and substrates by using support vector machines.J. Chem. Inf. Model. 45, 982-992 (2005)

### CYP2C9S - Substrate of CYP2C9
Cytochrome P450 enzymes are essential for the metabolism of many medications. Although this class has more than 50 enzymes, six of them metabolize 90 percent of drugs, with the two most significant enzymes being CYP3A4 and CYP2D6. This model predicts whether the ligand will likely be a substrate for CYP2C9 and will likely be metabolised by the enzyme, an important consideration for drug half-life, efficacy, and overall drug safety.
Dataset: 691 (143 Yes | 548 No)
- Yap CW., et al. Prediction of cytochrome P450 3A4, 2D6, and 2C9 inhibitors and substrates by using support vector machines.J. Chem. Inf. Model. 45, 982-992 (2005)

### CYP2D6I - Inhibitor of CYP2D6

Cytochrome P450 enzymes are essential for the metabolism of many medications. Although this class has more than 50 enzymes, six of them metabolize 90 percent of drugs, with the two most significant enzymes being CYP3A4 and CYP2D6. This model predicts whether the ligand will likely inhibit the activity of CYP2D6, an important consideration for drug-contraindications, efficacy, and overall drug safety.

Dataset: 691 (178 Yes | 513 No)
- Yap CW., et al. Prediction of cytochrome P450 3A4, 2D6, and 2C9 inhibitors and substrates by using support vector machines. J. Chem. Inf. Model. 45, 982-992 (2005)

### CYP2D6S - Substrate of CYP2D6

Cytochrome P450 enzymes are essential for the metabolism of many medications. Although this class has more than 50 enzymes, six of them metabolize 90 percent of drugs, with the two most significant enzymes being CYP3A4 and CYP2D6. This model predicts whether the ligand will likely be a substrate for CYP2D6 and will likely be metabolised by the enzyme, an important consideration for drug half-life, efficacy, and overall drug safety.

Dataset: 690 (195 Yes | 495 No)
- Yap CW., et al. Prediction of cytochrome P450 3A4, 2D6, and 2C9 inhibitors and substrates by using support vector machines. J. Chem. Inf. Model. 45, 982-992 (2005)

### CYP3A4I - Inhibitor of CYP3A4

Cytochrome P450 enzymes are essential for the metabolism of many medications. Although this class has more than 50 enzymes, six of them metabolize 90 percent of drugs, with the two most significant enzymes being CYP3A4 and CYP2D6. This model predicts whether the ligand will likely inhibit the activity of CYP2C9, an important consideration for drug-contraindications, efficacy, and overall drug safety.

Dataset: 690 (238 Yes | 452 No)
- Yap CW., et al. Prediction of cytochrome P450 3A4, 2D6, and 2C9 inhibitors and substrates by using support vector machines. J. Chem. Inf. Model. 45, 982-992 (2005)

### CYP3A4S - Substrate of CYP3A4

Cytochrome P450 enzymes are essential for the metabolism of many medications. Although this class has more than 50 enzymes, six of them metabolize 90 percent of drugs, with the two most significant enzymes being CYP3A4 and CYP2D6. This model predicts whether the ligand will likely be a substrate for CYP3A4 and and will likely be metabolised by the enzyme, an important consideration for drug half-life, efficacy, and overall drug safety.

Dataset: 691 (362 Yes | 329 No)
- Yap CW., et al. Prediction of cytochrome P450 3A4, 2D6, and 2C9 inhibitors and substrates by using support vector machines. J. Chem. Inf. Model. 45, 982-992 (2005)

*Toxicity:*
### AMES - A Mutagenicity Assay

The Ames test is an in vitro assay that utilizes bacteria to assay potential for inducing mutation within an organism's genetic material. The biological assays ability to assess the mutagenic potential of chemical compounds is a common method for evaluating potential drugs. This model predicts whether a query ligand will likely test positive for being mutagenic in an Ames test.

Dataset: 6505 (3496 Yes | 3009 No)

- Hansen K., et al. Benchmark Data Set for in Silico Prediction of Ames Mutagenicity. J. Chem. Inf. Model. 49, 2077-2081 (2009)

**AR - Androgen Receptor Agonist**

Androgen receptors are nuclear receptors that are activated by binding androgenic hormones, testosterone, or dihydrotestosterone and play a key biological role in regulating gene expression, particularly in the maintenance of the male sexual characteristics. Drug molecules that activate these receptors impacts both physical and mental health. This model will predict whether a query ligand will likely agonize androgen receptors, a critical consideration for drug safety.

Dataset: 5905 (224 Yes | 5681 No)

- Huang R., et al. Tox21 Challenge to Build Predictive Models of Nuclear Receptors and Stress Response Pathways as Mediated by Exposure to Environmental Toxicants and Drugs. Front. Environ. Sci. doi: 10.3389/978-2-88945-197-5 (2017)

**Carcin - Carcinogenic potency**

Carcinogenic potency is a metric used to quantify the cancer risk associated with exposure to a given chemical. Depending on the application, humans may be exposed to large volumes of small molecule drugs, making it critical to estimate carcinogenic potency to assess drug safety.

Dataset: 796 (415 Yes | 381 No)

- Fjodorova N., et al. Rodent Carcinogenicity Dataset. Dataset Papers in Medicine. 2013, 361615 (2013)

**ERa - Estrogen receptor 1 Agonist**

Estrogen receptor 1 is a nuclear receptor activated by the sex hormone estrogen and is responsible for maintaining the female sexual phenotype. Small molecules can potentially modulate this receptor, which would greatly impact human health, making evaluation of this property critical to establish drug safety. This model will predict whether a query ligand will act like an agonist to estrogen receptor 1.

Dataset: 3235 (2593 Yes | 642 No)

References:

- Ng HW., et al. Development and Validation of Decision Forest Model for Estrogen Receptor Binding Prediction of Chemicals Using Large Data Sets. Chem Res Toxicol. 28, 2343-2351 (2015)

**hPXR - Human pregnane X receptor agonist**

Human pregnane X is a promiscuous nuclear receptor that acts as a sensor for xenobiotic compounds. Some of PXR protein activators are steroid hormones, bile acids, and fat-soluble vitamins. This model will predict whether a query molecule will act like an agonist to human pregnane X receptor.
Dataset: 1709 (172 Yes | 1537 No)
References:
- AbdulHameed M. D. M., et al. Predicting Rat and Human Pregnane X Receptor Activators Using Bayesian Classification Models. Chem. Res. Toxicol. 29, 1729-1740 (2016)

*Regression Datasets:*

**ESOL**
ESOL is a molecular solubility dataset for 1128 compounds. Retrieved from Moleculenet.ai
Dataset: 1128 Compounds
- Delaney, John S. "ESOL: estimating aqueous solubility directly from molecular structure." Journal of chemical information and computer sciences 44.3 (2004): 1000-1005.
- Zhenqin Wu, Bharath Ramsundar, Evan N. Feinberg, Joseph Gomes, Caleb Geniesse, Aneesh S. Pappu, Karl Leswing, Vijay Pande, MoleculeNet: A Benchmark for Molecular Machine Learning, arXiv preprint, arXiv: 1703.00564, 2017.

**Freesolv**
FreeSolv is another common benchmark for regression models of small molecules. The dataset contains experimentally measured free energy of solvation for 642 compounds. Retrieved from Moleculenet.ai
Dataset: 642 Compounds
- Mobley, David L., and J. Peter Guthrie. "FreeSolv: a database of experimental and calculated hydration free energies, with input files." Journal of computer-aided molecular design 28.7 (2014): 711-720.
- Zhenqin Wu, Bharath Ramsundar, Evan N. Feinberg, Joseph Gomes, Caleb Geniesse, Aneesh S. Pappu, Karl Leswing, Vijay Pande, MoleculeNet: A Benchmark for Molecular Machine Learning, arXiv preprint, arXiv: 1703.00564, 2017.

**Lipholicity**
A dataset of 4200 curated, experimentally-derived, octanol/water distribution coefficients to describe molecular lipophilicity. Originally deposited in Chembl, but retrieved from Moleculenet.ai
Dataset: 4200 Compounds
- Hersey, A. ChEMBL Deposited Data Set - AZ dataset; 2015. https://doi.org/10.6019/chembl3301361
- Zhenqin Wu, Bharath Ramsundar, Evan N. Feinberg, Joseph Gomes, Caleb Geniesse, Aneesh S. Pappu, Karl Leswing, Vijay Pande, MoleculeNet: A Benchmark for Molecular Machine Learning, arXiv preprint, arXiv: 1703.00564, 2017.

**LogS**

The aqueous solubility, logS, of organic molecules play a large role in the expected ADME properties of a small molecule drug. This model uses experimental data that measured the solubility, S, of organic molecules at 20-25°C in mol/L.

Dataset: 1675

- Wang, J., et al. Development of reliable aqueous solubility models and their application in drug-like analysis. Journal of Chemical Information and Modeling. 47, 1395-1404, (2007)
- Hou, T., et al. ADME Evaluation in Drug Discovery. 4. Prediction of Aqueous Solubility Based on Atom Contribution Approach. Journal of Chemical Information and Computer Sciences. 44, 266-275, (2004)